# High Dimensional Bayesian Optimization Using Dropout

**Cheng Li, Sunil Gupta, Santu Rana, Vu Nguyen, Svetha Venkatesh, Alistair Shilton**
Centre for Pattern Recognition and Data Analytics (PRaDA), Deakin University, Australia
cheng.l@deakin.edu.au

## Abstract

Scaling Bayesian optimization to high dimensions is challenging task as the global optimization of high-dimensional acquisition function can be expensive and often infeasible. Existing methods depend either on limited "active" variables or the additive form of the objective function. We propose a new method for high-dimensional Bayesian optimization, that uses a dropout strategy to optimize only a subset of variables at each iteration. We derive theoretical bounds for the regret and show how it can inform the derivation of our algorithm. We demonstrate the efficacy of our algorithms for optimization on two benchmark functions and two real-world applications - training cascade classifiers and optimizing alloy composition.

## 1 Introduction

From mixture products (e.g. shampoos, alloys) to processes for mixture (e.g. heat treatments for alloys), the need to find the optimal values of control variables to achieve a target product lies at the heart of most industrial processes. The complexity arises because we do not know the mathematical relationship between the control variables and the target - it is a *Black-Box Function*. This exploration done through experimental optimization is a laborious process and limited by resource restrictions and cost.

The process of experimental optimization quickly hits limit as soon as the number of control variables are increased. For example, since the Bronze Age, fewer than 12 elements have been combined to make alloys. But the periodic table contains 97 naturally occurring elements. Only a tiny fraction of the target space has been explored because of the underlying complexity. By increasing the number of elements to just 15 in order to find a high-strength alloy with 3 mixing levels per element, the search space escalates to more than 14 million choices [Xue *et al.*, 2016]. Another illustrative problem is that wing configuration design of a high speed civil transport (HSCT) aircraft may include upto 26 variables to reach a targeted wing configuration [Koch *et al.*, 1999].

Bayesian optimization (BO) [Snoek *et al.*, 2012; Nguyen *et al.*, 2016] is a powerful technique to optimize expensive, black-box functions. A classical BO uses Gaussian process (GP) [Rasmussen and Williams, 2005] to model the mean and variance of the target function. As the function is expensive to interrogate, a surrogate function (or acquisition function) is constructed from the GP to trade-off exploitation (where mean is high) and exploration (where uncertainty is high). The next sample is determined by maximizing the acquisition function. Scaling BO methods to handle functions in high dimension presents two main challenges. Firstly, the number of observations required by the GP grows exponentially as input dimensions increase. This implies more experimental evaluations are required, often expensive and infeasible in real applications. Secondly, global optimization for high-dimensional acquisition functions is intrinsically a hard problem and can be prohibitively expensive to be feasible [Kandasamy *et al.*, 2015; Rana *et al.*, 2017].

Solutions have been proposed to tackle high-dimensional Bayesian optimization. Wang *et al.* [2013] projected the high-dimensional space into a low-dimensional subspace and then optimized the acquisition function in a low-dimensional subspace (REMBO). The assumption that only some dimensions ($d_e \ll D$) are effective is often restrictive. Qian *et al.* [2016] studied the case when all dimensions are effective, but many of them only have a small bounded effect by using sequential random embedding to reduce the embedding gap. These methods may not work if all dimensions in the high-dimensional function are similarly effective. The additive decomposition assumption is another solution for high-dimensional function analysis. Kandasamy *et al.* [2015] proposed the Add-GP-UCB model in which the objective function is assumed to be the sum of a set of low-dimensional functions with disjoint dimensions and then BO can be performed in the low-dimensional space. Add-GP-UCB allows the objective function to vary along the entire feature domain. Li *et al.* [2016] generalized the Add-GP-UCB by eliminating an axis-aligned representation. However, in practice it is difficult to know the decomposition of functions in advance, especially for non-separable functions. The most related one to our work is DSA [Ulmasov *et al.*, 2016], which reduces the number of variables at each iteration by PCA. There are two problems when using DSA: (1) using PCA for selecting variables is effective only when there are large number of data points. This is especially not true for Bayesian optimization where in the beginning we do not have many points. Eigen vector estimation using small number of data points are often

Proceedings of the Twenty-Sixth International Joint Conference on Artificial Intelligence (IJCAI-17)inaccurate and can be misleading. (2) DSA may get stuck in a local optimum since it only clamps the other coordinates to their current best values.

This paper proposes an alternative approach that does not rely on the assumptions that the objective function depends on limited "active" features - projections can be made into lower dimensional sub-spaces (fixed [Djolonga *et al.*, 2013; Wang *et al.*, 2013] or updated [Ulmasov *et al.*, 2016]) or objective function decomposed in additive forms [Kandasamy *et al.*, 2015; Li *et al.*, 2016]. Motivated by the dropout algorithm in neural networks [Srivastava *et al.*, 2014], we explore dimension dropout in high-dimensional Bayesian optimization. We choose $d$ out of $D$ dimensions ($d < D$) randomly at each iteration and only optimize variables from the chosen dimensions via Bayesian optimization. To "fill-in" the variables from the left-out dimensions, we consider alternate strategies - random values, values of these variables from the best found function value thus far, and a mixture of these two methods. We formulate our dropout algorithms and apply them on benchmark functions and two real-world applications of training cascade classifiers [Viola and Jones, 2001] and an aluminium alloy design. We compare them with baselines ( random search, standard BO, REMBO [Wang *et al.*, 2013] and Add-GP-UCB [Kandasamy *et al.*, 2015] ). The experimental results demonstrate the effectiveness of our algorithms. We derive regret bound theoretically. As expected the cost of Dropout algorithm is a remaining "regret gap", and we provide insights to how this gap can be reduced through the strategies we have formulated to fill-in the dropped-out variables. Our main contributions are:

- Formulation of a novel variable dropout method for high-dimensional Bayesian optimization;
- Theoretical analysis of the regret bound for our dropout algorithm, and the use of the regret bound to guide how to fill-in the dropped out variables;
- Demonstration and comparison of our algorithms with baselines on two synthetic functions and two real applications: training cascade classifiers and designing an aluminium alloy through improved (phase) utility.

## 2 Formulation

**Preliminaries** Bayesian optimization is used to maximize or minimize a function $f$ in the input domain $\mathcal{X} \subseteq \mathbb{R}^D$. It includes two critical components: prior and acquisition functions. Gaussian process (GP) is a popular choice for the prior due to its tractability for posterior and predictive distributions and it is specified by its mean $m(.)$ and covariance kernel function $k(.,.)$. Give a set of observations $x_{1:t}$ and the corresponding values $\mathbf{f}(x_{1:t})$, the probability of any finite set of $\mathbf{f}$ is Gaussian

$$\mathbf{f}(\boldsymbol{x}) \sim \mathcal{N}(\mathbf{m}(\boldsymbol{x}), \mathbf{K}(\boldsymbol{x}, \boldsymbol{x}^{'})) \quad (1)$$

where $\mathbf{K}(\boldsymbol{x}, \boldsymbol{x}^{'})_{i,j} = k(\boldsymbol{x}_i, \boldsymbol{x}^{'}_j)$ is the covariance matrix. Two popular choice of $k$ are the squared exponential (SE) kernel and the Matérn kernel. The predictive distribution of a new point $x_{t+1}$ is given as
$f_{t+1} \mid \mathbf{f}_{1:t} \sim \mathcal{N}(\mu_{t+1}(\boldsymbol{x}_{t+1} \mid \boldsymbol{x}_{1:t}, \mathbf{f}_{1:t}), \sigma^2_{t+1}(\boldsymbol{x}_{t+1} \mid \boldsymbol{x}_{1:t}, \mathbf{f}_{1:t}))$
$\qquad (2)$

where $\mathbf{f}_{1:t} = \mathbf{f}(\boldsymbol{x}_{1:t})$, $\mu_{t+1}(.) = \mathbf{k}^T \mathbf{K}^{-1} \mathbf{f}_{1:t}$, $\sigma^2_{t+1}(.) = k(\boldsymbol{x}_{t+1}, \boldsymbol{x}_{t+1}) - \mathbf{k}^T \mathbf{K}^{-1} \mathbf{k}$ and $\mathbf{k} = [k(\boldsymbol{x}_{t+1}, \boldsymbol{x}_1), \cdots, k(\boldsymbol{x}_{t+1}, \boldsymbol{x}_t)]$.

Acquisition function is a proxy function derived from the predictive mean and variance and it determines the next sample point. We denote acquisition function $a(\boldsymbol{x} \mid \{\boldsymbol{x}_{1:t}, f_{1:t}\})$ and the next sample point $\boldsymbol{x}_{t+1} = \text{argmax}_{\boldsymbol{x} \in \mathcal{X}} a(\boldsymbol{x} \mid \{\boldsymbol{x}_{1:t}, f_{1:t}\})$. Some examples of acquisition functions include Expected Improvement (EI) and GP-UCB [Srinivas *et al.*, 2010]. The EI-based acquisition function is to compute the expected improvement with respect to the current maximum $f(\boldsymbol{x}^+)$, or $\mathbf{EI}(\boldsymbol{x}) = \mathbb{E}(\max\{0, f_{t+1}(\boldsymbol{x}) - f(\boldsymbol{x}^+)\} \mid \boldsymbol{x}_{1:t}, \mathbf{f}_{1:t})$. The closed form has been derived in [Mockus *et al.*, 1978; Jones *et al.*, 1993]. The GP-UCB [Srinivas *et al.*, 2010] is defined as $\text{UCB}(\boldsymbol{x}) = \mu(\boldsymbol{x}) + \sqrt{\beta}\sigma(\boldsymbol{x})$, where $\beta$ is a positive tradeoff. The first term contributes to the exploitation and the second term contributes to the exploration. DIRECT [Jones *et al.*, 1993] is often used to find the global maximum in acquisition function.

We seek to maximize a function $f$ in the restricted domain $\mathcal{X} = [0, 1]^D$ (this can always be achieved by scaling). We assume the maximal function value can be achieved at a query point $\boldsymbol{x}^*$, i.e. $\boldsymbol{x}^* = \text{argmax}_{\boldsymbol{x} \in \mathcal{X}} f(\boldsymbol{x})$. At iteration $t$ and the corresponding query point $\boldsymbol{x}_t \in \mathcal{X}$, the instantaneous regret $r_t$ is defined $r_t = f(\boldsymbol{x}^*) - f(\boldsymbol{x}_t)$ and the cumulative regret $R_T$ is defined $R_T = \sum_{t=1}^{T} r_t$. A desirable property of an algorithm is to have no-regret: $\lim_{T \to \infty} \frac{1}{T} R_T = 0$.

**Dropout Algorithms** We refer to $I_d$ as the indices of $d$ out of $D$ dimensions and $I_{D-d}$ as the indices of the left-out $D - d$ dimensions so that $I_d \bigcup I_{D-d} = \{1, \cdots, D\}$ and $I_{D-d} \cap I_d = $ . The corresponding variables from $I_d$ and $I_{D-d}$ dimensions are respectively denoted as $\boldsymbol{x}^{I_d}$ and $\boldsymbol{x}^{I_{D-d}}$. For the convenience we later use $\boldsymbol{x}^d = \boldsymbol{x}^{I_d}$, $\boldsymbol{x}^{D-d} = \boldsymbol{x}^{I_{D-d}}$ and hence $\boldsymbol{x} = [\boldsymbol{x}^d, \boldsymbol{x}^{D-d}]$.

Motivated by the dropout algorithm in neural networks [Srivastava *et al.*, 2014], we explore dimension dropout for high-dimensional Bayesian optimization. We randomly choose $d$ out of $D$ dimensions ($d < D$) at each iteration and only optimize the $d$-dimensional variables through Bayesian optimization. Specifically we assume in the $d$-dimensional space the observations $y = f([\boldsymbol{x}^d, \boldsymbol{x}^{D-d}]) + \varepsilon$ with $\varepsilon \sim \mathcal{N}(0, \sigma^2)$ for all $\boldsymbol{x}^{D-d}$. Gaussian process then is used to model the function values $f([\boldsymbol{x}^d, \boldsymbol{x}^{D-d}]), \forall \boldsymbol{x}^{D-d}$. The predictive mean $\mu(\boldsymbol{x}^d)$ and variance $\sigma(\boldsymbol{x}^d)$ can be computed. As with GP-UCB [Srinivas *et al.*, 2010], we construct the acquisition function in the $d$-dimensional space

$$a(\boldsymbol{x}^d) = \mu_{t-1}(\boldsymbol{x}^d) + \sqrt{\beta^d_t}\sigma_{t-1}(\boldsymbol{x}^d) \quad (3)$$

where $\beta^d_t$ is a factor controlling tradeoff between the exploitation $\mu_{t-1}(\boldsymbol{x}^d)$ and exploration $\sigma_{t-1}(\boldsymbol{x}^d)$. At each iteration, we determine a new $\boldsymbol{x}^d$ by maximizing the Eq .(3).

Given $\boldsymbol{x}^d_t$, we still need to fill-in the variables $\boldsymbol{x}^{D-d}_t$ from the left-out $D - d$ dimensions to evaluate the function. We **devise three "fill-in" strategies** for $\boldsymbol{x}^{D-d}_t$:

- **Dropout-Random**: use a random value in the domain:

$$\boldsymbol{x}^{D-d}_t \sim u(\boldsymbol{x}^{D-d}) \quad (4)$$

where $u(\cdot)$ is a uniform distribution.

- **Dropout-Copy:** copy the value of the variables from the best function value so far:

$$\boldsymbol{x}_t^+ = \mathrm{argmax}_{t' \leq t} f(\boldsymbol{x}_{t'})$$
$$\boldsymbol{x}_t^{D-d} = (\boldsymbol{x}_t^+)^{D-d} \quad (5)$$

where $\boldsymbol{x}_t^+$ is the variables of the best found function value till $t$ iterations.

- **Dropout-Mix:** use a mixture of the above two methods. We use a random value with probability $p$ or copy the value from the variables of the best found function value so far with the probability $1-p$.

Dropout-Random does not work effectively when a large number of dimensions are influential. However, it can still improve the optimization since we optimize $d$ variables each iteration. Copying the value of the variables from the best function value so far seems to be an efficient strategy to consistently improve the previous best regret. However, this method may get stuck in a local optimum. This problem is solved by the third strategy, which helps the copy method to escape the local optimum with a probability $p$.

Our approach performs Bayesian optimization in the $d$-dimensional space and thus DIRECT requires $\mathcal{O}(\zeta^{-d})$ calls to the acquisition function to achieve $\zeta$ accuracy [Jones *et al.*, 1998]. It is significantly better than full-dimensional BO where DIRECT requires $\mathcal{O}(\zeta^{-D})$ objective function calls. Both our approach and the full-dimensional BO need the time complexity $\mathcal{O}(n^3)$ to compute the inverse of the covariance matrix, where $n$ is the number of observations. We summarize our algorithms in the following.

---

**Algorithm 1** Dropout Algorithm for High-dimensional Bayesian Optimization

---

**Input:** $\mathcal{D}_1 = \{\boldsymbol{x}_0, y_0\}$
1: **for** $t = 1, 2, \cdots$ **do**
2:    randomly select $d$ dimensions
3:    $\boldsymbol{x}_t^d \leftarrow \mathrm{argmax}_{\boldsymbol{x}_t^d \in \mathcal{X}^d} a(\boldsymbol{x}^d \mid \mathcal{D}_t)$ (Eq.(3))
4:    $\boldsymbol{x}_t^{D-d} \leftarrow$ one of three "fill-in" strategies (Sec 2.)
5:    $\boldsymbol{x}_t \leftarrow \boldsymbol{x}_t^d \cup \boldsymbol{x}_t^{D-d}$
6:    $y_t \leftarrow$ Query $y_t$ at $\boldsymbol{x}_t$
7:    $\mathcal{D}_{t+1} = \mathcal{D}_t \cup \{\boldsymbol{x}_t, y_t\}$
8: **end for**

---

## 3 Theoretical Analysis

Our main contribution is to derive a regret bound for our algorithm and discuss heuristic strategies. We denote $f(\boldsymbol{x}^d)$ as the worst function value given $\boldsymbol{x}^d$, i.e. $f(\boldsymbol{x}^d) = f([\boldsymbol{x}^d, \boldsymbol{x}_w^{D-d}])$, where $\boldsymbol{x}_w^{D-d} = \mathrm{argmin}_{\boldsymbol{x}^{D-d}} f([\boldsymbol{x}^d, \boldsymbol{x}^{D-d}])$.

**Assumption 1.** *Let $f$ sample from GP with the kernel $k(\boldsymbol{x}, \boldsymbol{x}')$, which is L-Lipschitz for all $\boldsymbol{x}$. Then the partial derivatives of $f$ satisfy the following high probability bound for some constants $a, b > 0$,*

$$\mathcal{P}\left(\forall j, \frac{\partial f}{\partial x_j} < L\right) \geq 1 - ae^{-(L/b)^2}, \forall t \geq 1 \quad (6)$$

The assumption implies the following equation holds with the probability greater than $1 - \delta/2$ for all $\boldsymbol{x}$,

$$\mid f(\boldsymbol{x}) - f(\boldsymbol{x}^d) \mid = \mid f([\boldsymbol{x}^d, \boldsymbol{x}^{D-d}]) - f([\boldsymbol{x}^d, \boldsymbol{x}_w^{D-d}]) \mid$$
$$\leq L \|\boldsymbol{x}^{D-d} - \boldsymbol{x}_w^{D-d}\|_1 \leq L|D-d| \quad (7)$$

where $L = b\sqrt{\log(2(D-d)a/\delta)}$.

**Lemma 2.** *Pick $\delta \in (0,1)$ and set $\beta_t^d = 2\log(4\pi_t/\delta) + 2d \log\left(dt^2 br \sqrt{\log(4da/\delta)}\right)$, where $\Sigma_{t \geq 1} \pi_t^{-1} = 1, \pi_t > 0$. Then in the $d$-dimensional space, with the probability $\geq 1 - \delta/2$*

$$\mid f(\boldsymbol{x}^d) - \mu_{t-1}(\boldsymbol{x}^d) \mid \leq \sqrt{\beta_t^d} \sigma_{t-1}(\boldsymbol{x}^d), \forall t \geq 1 \quad (8)$$

The Lemma 2 is derived from the Bayesian optimization in the $d$-dimensional space. The proof is identical to Theorem 2 in [Srinivas *et al.*, 2010].

**Lemma 3.** *Let $\beta_t^d$ be defined as in Lemma 2 and set $\sigma'_{t-1}(\boldsymbol{x}^d) = \sigma_{t-1}(\boldsymbol{x}^d) + \frac{L(D-d)}{\sqrt{\beta_t^d}}$. Then*

$$\mid f(\boldsymbol{x}) - \mu_{t-1}(\boldsymbol{x}^d) \mid \leq \sqrt{\beta_t^d} \sigma'_{t-1}(\boldsymbol{x}^d) \quad (9)$$

*holds with the probability $\geq 1 - \delta$.*

Proof. The following is true for all $t \geq 1$ and for all $\boldsymbol{x} \in \mathcal{X}$ with probability $> 1 - \delta$,

$$\mid f(\boldsymbol{x}) - \mu_{t-1}(\boldsymbol{x}^d) \mid$$
$$\leq \mid f(\boldsymbol{x}) - f(\boldsymbol{x}^d) \mid + \mid f(\boldsymbol{x}^d) - \mu_{t-1}(\boldsymbol{x}^d) \mid$$
$$\leq L\|\boldsymbol{x} - [\boldsymbol{x}^d, \boldsymbol{x}_w^{D-d}]\|_1 + \sqrt{\beta_t^d} \sigma_{t-1}(\boldsymbol{x}^d) \quad (10)$$
$$\leq L(D-d) + \sqrt{\beta_t^d} \sigma_{t-1}(\boldsymbol{x}^d)$$
$$\leq \sqrt{\beta_t^d} \sigma'_{t-1}(\boldsymbol{x}^d)$$

where $\sigma'_{t-1}(\boldsymbol{x}^d) = \sigma_{t-1}(\boldsymbol{x}^d) + \frac{L(D-d)}{\sqrt{\beta_t^d}}$. The Line 3 to Line 4 In Eq.(10) exploits Eq.(7). The variance difference $\frac{L(D-d)}{\sqrt{\beta_t^d}}$ will reduce with iteration $t$ since $\beta_t^d$ is increasing.

**Lemma 4.** *Pick $\delta \in (0,1)$ and Let $\beta_t^d$ be defined as in Lemma 2. Then the regret $r_t$ is bounded by $2\sqrt{\beta_t^d} \sigma'_{t-1}(\boldsymbol{x}^d) + \frac{1}{t^2}$.*

Proof. By definition of $\boldsymbol{x}_t^d$: $\mu_{t-1}(\boldsymbol{x}_t^d) + \sqrt{\beta_t^d} \sigma'_{t-1}(\boldsymbol{x}_t^d) \geq \mu_{t-1}([\boldsymbol{x}^*]_t^d) + \sqrt{\beta_t^d} \sigma'_{t-1}([\boldsymbol{x}^*]_t^d)$. According to Lemma 5.7 in [Srinivas *et al.*, 2010], $\mu_{t-1}([\boldsymbol{x}^*]_t^d) + \sqrt{\beta_t^d} \sigma'_{t-1}([\boldsymbol{x}^*]_t^d) + \frac{1}{t^2} \geq f(\boldsymbol{x}^*)$, where $[\boldsymbol{x}^*]_t$ denotes the closest point in discretizations $D_t \subset \mathcal{X}$ to $\boldsymbol{x}^*$. Then

$$r_t = f(\boldsymbol{x}^*) - f(\boldsymbol{x}_t)$$
$$\leq \sqrt{\beta_t^d} \sigma'_{t-1}(\boldsymbol{x}_t^d) + \mu_{t-1}(\boldsymbol{x}_t^d) + \frac{1}{t^2} - f([\boldsymbol{x}_t^d, \boldsymbol{x}_t^{D-d}])$$
$$\leq 2\sqrt{\beta_t^d} \sigma'_{t-1}(\boldsymbol{x}^d) + \frac{1}{t^2}$$

**Lemma 5.** *Pick $\delta \in (0,1)$ and let $\beta_t^d$ be defined as Lemma 2. Then the cumulative regret holds with the probability $\geq 1 - \delta$ and $C_1 = 8/\log(1 + \sigma^{-2})$,*

$$R_T \leq \sqrt{C_1 \beta_T^d \gamma_T T} + 2TL(D-d) + 2 \quad (11)$$

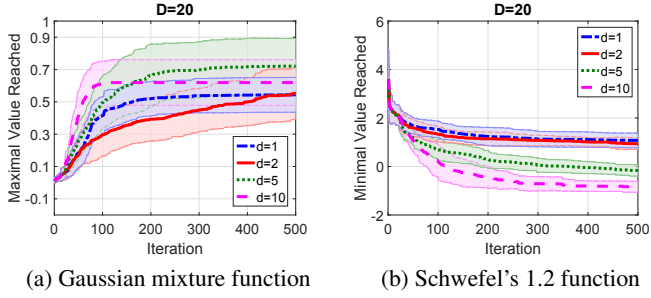

(a) Gaussian mixture function  (b) Schwefel's 1.2 function

Figure 1: The effect of the number of dimensions $d$ in Dropout-Copy. The y-axis in Gaussian mixture function presents the real function value (Higher value is better). The y-axis in Schwefel's 1.2 function presents the logarithm of function value (Lower value is better).

Proof. we prove Lemma 5 in the following

$$R_T = \sum_{t \leq T} r_t \leq \sum_{t \leq T} \left( 2\sqrt{\beta_t^d} \sigma'_{t-1}(\boldsymbol{x}^d) + \frac{1}{t^2} \right)$$

$$= \sum_{t \leq T} \left( 2\sqrt{\beta_t^d} \sigma_{t-1}(\boldsymbol{x}^d) \right) + 2TL(D-d) + \frac{\pi^2}{6} \quad (12)$$

$$\leq \sqrt{C_1 \beta_T^d \gamma_T T} + 2TL(D-d) + 2$$

where $\sum_{t \leq T} \left( 2\sqrt{\beta_t^d} \sigma_{t-1}(\boldsymbol{x}^d) \right)$ in the second line of Eq.(12) can be bounded via the Theorem 1 in [Srinivas *et al.*, 2010]. $\gamma_T$ can be bounded for different kernels. For the SE kernel, $\gamma_T = \mathcal{O}((\log T)^{d+1})$.

### Discussion

Lemma 5 indicates $\lim_{T \to \infty} \frac{1}{T} R_T \leq 2L(D-d)$ - that is, a regret gap remains in the limit. This is introduced through the bound on the worst-case of $||\boldsymbol{x} - [\boldsymbol{x}^d, \boldsymbol{x}^{D-d}]||_1, \forall \boldsymbol{x}^{D-d}$ (Eq.(7)). In reality a judicious choice of the $D-d$ dimensions will improve this bound by reducing this difference. We have thus formulated three options for filling in the variables of the dropped out dimensions: random, best value and mixture of the two. Intuitively if the current optimum is far away from the global optimum, random values are an appropriate guess for the "fill-in" as there is no other information. If the current optimum is close to the global optimum, copying the value of the dropped-out variables from the best found function value is likely to improve the best regret obtained by the previous iterations. This behaves like block coordinate descent [Nesterov, 2012] that optimizes a chosen block of coordinates while keeping others fixed. The difference is that block coordinate descent assumes that the previous iteration has reached the best value, while Dropout-Copy starts from the best of previous iterations. When the current optimum value is close to a local optimum, Dropout-Copy may get stuck. To escape this local optimum, we use Dropout-Mix that introduces a random fill-in with a pre-specified probability into Dropout-Copy.

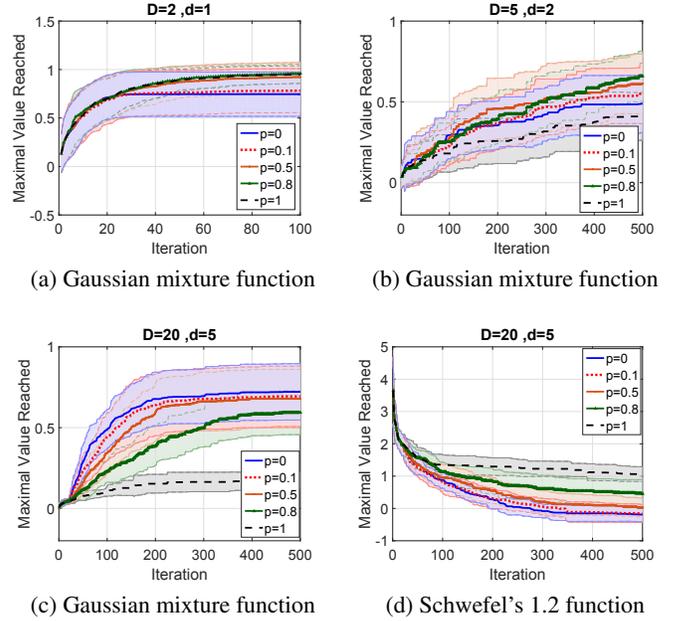

(a) Gaussian mixture function  (b) Gaussian mixture function

(c) Gaussian mixture function  (d) Schwefel's 1.2 function

Figure 2: The effect of the probability $p$ in Dropout-Mix.

## 4 Experiments

We evaluate our methods on benchmark functions and two real applications. We compare our methods with four baselines: random search, which is a simple random sampling method, the standard BO, REMBO [Wang *et al.*, 2013] which projects high dimensions to lower dimensions, and ADD-GP-UCB [Kandasamy *et al.*, 2015] which optimizes disjoint groups of variables and combines them. For standard BO we allocate a budget of 30 seconds (which is larger than the time returned by our algorithms) to optimize the acquisition function at each iteration. The number of initial observations are set at $d+1$. We use the SE kernel with the lengthscale 0.1 and DIRECT [Jones *et al.*, 1993] to optimize acquisition functions. We run each algorithm 20 times with different initializations and report the average value with standard error.

### 4.1 Optimization of Benchmark Functions

To demonstrate that Dropout-Mix can deal with local convergence, we choose a bimodal Gaussian mixture function as our test function. The Gaussian mixture function is defined as $y = \mathcal{NP}(\boldsymbol{x}; \mu_1, \Sigma_1) + \frac{1}{2} \mathcal{NP}(\boldsymbol{x}; \mu_2, \Sigma_2)$, where $\mathcal{NP}$ is the Gaussian probability function, $\mu_1 = [2, 2, \cdots, 2]$, $\mu_2 = [3, 3, \cdots, 3]$ and $\Sigma_1 = \Sigma_2 = diag([1, 1, \cdots, 1])$. The domain of definition $\mathcal{X} = [1, 4]^D$ and its global maximum is located at $\boldsymbol{x}^* = [2, 2, \cdots 2]$. This function has a local maximum and no interacting variables. To demonstrate that our algorithms can effectively work for functions with interacting variables, we use the unimodal Schwefel's 1.2 function $f(x) = \sum_{j=1}^{D} \left( \sum_{i=1}^{j} \boldsymbol{x}_i \right)^2$ as our second test function. It is defined in the domain $\mathcal{X} = [-1, 1]^D$ and has the global minimum at $\boldsymbol{x}^* = [0, 0, \cdots 0]$. We compare the algorithms in terms of the best function values reached upto any iteration.

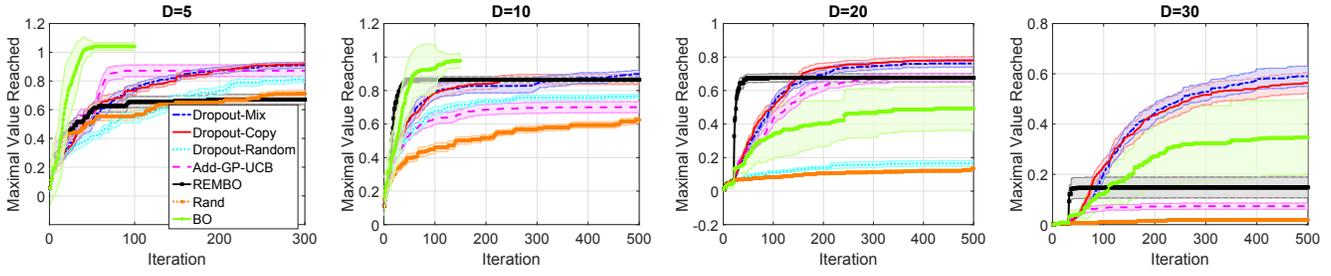

Figure 3: The optimization for the Gaussian mixture function. Higher value is better. Four different dimensions are tested from left to right (a) $D = 5$ (b) $D = 10$ (c) $D = 20$ (d) $D = 30$. The BO for $D = 5$ and $D = 10$ is terminated once it converges. The graphs are best seen in color.

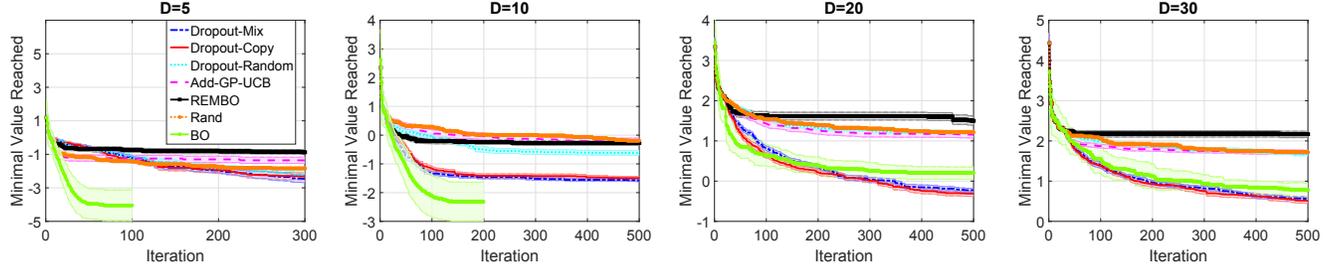

Figure 4: The optimization for Schwefel's 1.2 function. Lower value is better. Four different dimensions are tested from left to right (a) $D = 5$ (b) $D = 10$ (c) $D = 20$ (d) $D = 30$. The graphs are best seen in color.

**On the number of chosen dimensions** $d$   We investigate how the number of chosen dimensions $d$ affects the dropout algorithms. We experiment $d = 1, 2, 5, 10$ for $D = 20$ in Dropout-Copy. For one experiment, we keep the same number of dimensions for all iterations. The results for two test functions are shown in Figure 1. Since the optimized Gaussian mixture function has non-interacting variables, then variables can be optimized independently so that Dropout-Copy with $d = 1$ can still improve the maximal value reached. Dropout-Copy with $d = 5$ performs best. However, for the Schwefel's 1.2 function independently optimizing variables is not an efficient way. Figure 1 (b) shows that there is a faster convergence rate for a larger $d$. It can be explained by noting that a large $d$ has relatively higher probability of optimizing interacting variables within stipulated iterations. These graphs show that it is reasonable to compromise: if $d$ is large then global optimization in the $d$-dimensional space might be costly, and if $d$ is small then the convergence rate is slow for functions with interacting variables.

**On the probability** $p$   To study the influence of the probability $p$ in Dropout-Mix, we set $p = 0, 0.1, 0.5, 0.8, 1$. Dropout-Copy and Dropout-Random are the special cases of Dropout-Mix ($p = 0$ and $p = 1$). In this experiment, for the low-dimensional Gaussian mixture functions ($D = 2$ and $D = 5$), we set $\mu_1 = [2, 2, \cdots, 2]$, $\mu_2 = [5, 5, \cdots, 5]$, $\mathcal{X} = [0, 7]^D$ so that two modes are far. We use $d = 1$ for $D = 2$ and $d = 2$ for $D = 5$. For both high-dimensional Gaussian mixture functions and Schwefel's function, we keep the same setting as before and use $d = 5$ for $D = 20$. The results are shown in Figure 2 (a), (b) and (c).

We see that Dropout-Mix and Dropout-Random work in low dimensions. Dropout-Copy does not work well and because it may get stuck in a local optimum for low-dimensional functions. This happens with a lower probability in high dimensions and thus the average performance of Drop-Copy is slightly better than Dropout-Mix in the Figure 2 (c). Dropout-Copy always performs best for the unimodal function, seen in Figure 2 (d). Therefore, we recommend using Drop-Copy and Drop-Mix with a small $p$ (e.g. 0.1, 0.2) in high-dimensions.

**Comparison with existing approaches**   Based on the experiments above, we test our algorithms with $d = 2$ for $D = 5$ and $d = 5$ for $D = 10, 20, 30$. Dropout-mix is applied with $p = 0.1$. For the Gaussian mixture function, we set $\mu_1 = [2, 2, \cdots, 2]$ and $\mu_2 = [3, 3, \cdots, 3]$ for all $D$. Since we do not know the structure of the functions, we use REMBO with a $d$-dimensional projection and ADD-GP-UCB with $d$ dimensions at each group, where the value $d$ is the same with dropout algorithms. We run 500 function evaluations for these two functions. The results are shown in Figure 3 and 4 respectively. In low dimensions ($D = 5$ and $D = 10$), standard BO performs best. In high dimensions ($D = 20$ and $D = 30$), our Dropout-Mix and Dropout-Copy significantly outperform other baselines. Not surprisingly, REMBO and ADD-GP-UCB do not perform well since the intrinsic structure of functions does not fit their prior assumptions.

### 4.2 Training Cascade Classifier

We evaluate the dropout algorithm by training a cascade classifier [Viola and Jones, 2001] on three real datasets from UCI

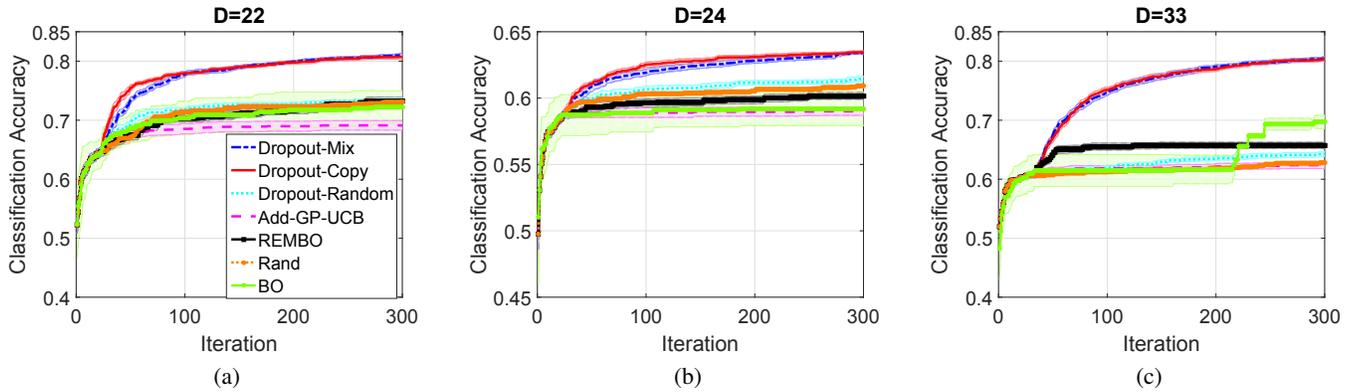

Figure 5: Maximum classification accuracy for training data as a function of Bayesian optimization iteration. The number of stages in a cascade classifier is equal to the number of features in three datasets (a) IJCNN1 $D = 22$, (b) German $D = 24$, (c) Ionosphere $D = 33$.

repository: IJCNN1, German and Ionosphere dataset. A cascade classifier consists of a series of weak classifiers. Each weak classifier is a simple decision stump. The weight for instances is updated based on the error rate from the previous weak classifier. Therefore, the threshold at the decision stump is very important. Generally, independently computing the thresholds is not an optimal strategy. We seek to find optimal thresholds by maximizing the training accuracy. Features in all datesets are scaled between [0, 1]. The number of stages is set equal to the number of features in the dataset. We use $d = 5$ and $p = 0.1$ for all datasets. We ensure that the dimension at each group in Add-GP-UCB is lower than 10 so that DIRECT can work. The experimental results are shown in Figure 5. Dropout-Copy and Dropout-Mix perform similarly but significantly better than other methods.

### 4.3 Alloy Design

AA-2050 is a low density high corrosion resistant alloy and is used for aerospace applications. The current alloy has been designed decades ago and is considered by our metallurgist collaborator as a prime candidate for further improvement. To measure utility of an alloy composition we use the software based thermodynamic simulator (THEMOCALC). The alloy consists of 9 elements (Al, Cu, Mg, Zn, Cr, Mn, Ti, Zr, and Sc). The utility is defined by a weighted combination of four phases that are produced. The phases relate to the internal structures and influence alloy properties. In all we have a 13-dimensional optimization problem (9 elements and 4 operational parameters). We seek the composition that maximizes this utility. The result is given in Figure 6. We started from a utility of about 4.5. After 500 optimization iterations, our Dropout-Mix can achieve the utility of 5.1 while the standard BO keeps around 4.8 after 100 iterations. The results clearly show the effectiveness of our methods for the real world application of alloy design.

## 5 Conclusion and Future Work

We propose a new method for high-dimensional Bayesian optimization by using a drop-out strategy. We develop three strategies to fill-in the variables from the dropped-out dimensions, including random values, the value from the best found sample so far and the mixture of these methods. The regret bounds for our methods has been derived and discussed. Our experimental results on synthetic and real applications show that our methods works effectively for the high-dimensional optimization. It might be promising if we only apply local optimization to the acquisition function built from the dropped-out dimensions. We intend to consider more efficient ways to choose dimensions in future.

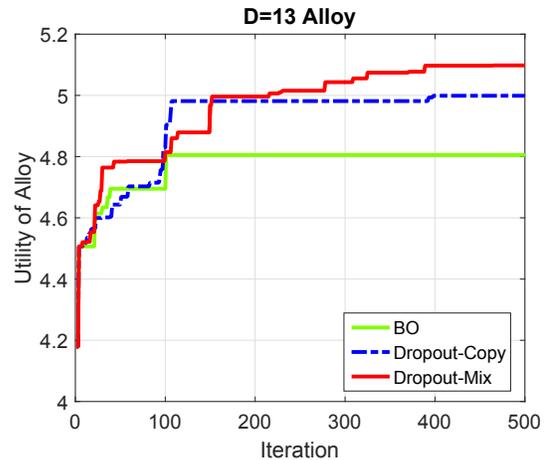

Figure 6: The utility of alloy vs the iterations of Bayesian optimization. The number of optimal parameters is 13. We use $d = 5$ in this experiment.

## Acknowledgments

This work is partially supported by the Telstra-Deakin Centre of Excellence in Big Data and Machine Learning. We thank anonymous reviewers for their valuable comments.